\documentclass[runningheads]{llncs}
\usepackage{graphicx}
\usepackage{xcolor}
\usepackage{color}
\usepackage{amssymb}

\usepackage{amsmath}
\newcommand\myeq{\mathrel{\overset{\makebox[0pt]{\mbox{\normalfont\tiny\sffamily def}}}{=}}}

\usepackage{array}
\newcolumntype{P}[1]{>{\centering\arraybackslash}p{#1}}

\begin{document}
\title{What am I allowed to do here?: Online Learning of Context-Specific Norms by Pepper}%\thanks{This work was partially supported by the Air Force Office of Sponsored Research contract FA9550-17-1-0017 and National Science Foundation grant CNS-1830390. Any opinions, findings, and conclusions or recommendations expressed in this material are those of the author(s) and do not necessarily reflect the views of the National Science Foundation.}}
\titlerunning{Online Learning of Context-Specific Norms by Pepper}
% If the paper title is too long for the running head, you can set
% an abbreviated paper title here
%
%\inst{1}
\author{Ali Ayub%\orcidID{0000-0001-9458-477X} \and
\and 
Alan R. Wagner}
\authorrunning{Ayub, A. and Wagner, A. R.}
% First names are abbreviated in the running head.
% If there are more than two authors, 'et al.' is used.
%
\institute{The Pennsylvania State University, State College, PA, 16802, USA \\
\email{\{aja5755,alan.r.wagner\}@psu.edu}}
\maketitle              % typeset the header of the contribution
\begin{abstract}
    Social norms support coordination and cooperation in society. With social robots becoming increasingly involved in our society, they also need to follow the social norms of the society. This paper presents a computational framework for learning contexts and the social norms present in a context in an online manner on a robot. The paper utilizes a recent state-of-the-art approach for incremental learning and adapts it for online learning of scenes (contexts). The paper further utilizes Dempster-Schafer theory to model context-specific norms. After learning the scenes (contexts), we use active learning to learn related norms. We test our approach on the Pepper robot by taking it through different scene locations. Our results show that Pepper can learn different scenes and related norms simply by communicating with a human partner in an online manner\footnote{The final authenticated publication is available online at https://doi.org/10.1007/978-3-030-62056-1\_19}. 
    
\end{abstract}

\section{Introduction}
\label{sec:intro}

Norms are an implicit part of any society which guide the actions taken by humans in that society. Norms support the actions taken by humans across time and generations and bring coordination and cooperation in a society \cite{ullmann76}. Some norms are defined explicitly in the form of laws. Social norms, however, are more implicit and are learned through the actions of the community. For example, when in a library, people can implicitly learn that talking is impermissible, while at a party talking is not only permissible but obligatory. %Humans learn these context-specific norms 

With robots increasingly becoming an integral part of the society in different roles, such as household robots or socially assistive robots, it is necessary that they follow the implicit social norms of the society. Since social norms can differ based on different communities, it is impossible to pre-program a universal set of norms. Thus, similar to humans, robots must learn norms in a lifelong manner using different sources, such as by asking questions or observing other humans. The ultimate goal of this paper is to develop a method allowing a robot to learn different scenes and the norms associated with the scene while it wanders around different environments.

Despite its importance for social robotics applications, work in context-specific norm learning on robots has been limited. Sarathay et al. \cite{Sarathy17_IEEE,Sarathy17} present a method for formal modeling of context-specific norms, however they do not apply their approach on a real robot. For online learning of norms on a robot, it is first necessary to develop perceptual systems that allow the robot to learn different scene categories. The term scene is used here to describe the perception-level representation of a context. To the best of our knowledge, we are among the first to continually learn both the context a robot is operating in and its associated acceptable behaviors using only streaming video data.

In this paper, we present a computational framework to learn different scene categories and associated norms from streaming video data in an online manner. Online learning is a field of machine learning in which a single system is required to both incrementally learn and perform open-set recognition. Where, incremental learning is a field of machine learning focused on creating systems that can incrementally learn over time without requiring all the training data to be available in a single batch \cite{Ayub_2020_CVPR_Workshops,Ayub_ICML_20,Rebuffi_2017_CVPR,Wu_2019_CVPR,Ayub_RoMan_20}. Open-set recognition, on the other hand, uses a trained model to predict if a new instance of data belongs to an already learned class or is contains entirely new items. Thus making the online learning problem extremely challenging. We adapt a recent state-of-the-art incremental learning approach termed Centroid-Based Concept Learning and Pseudo-Rehearsal (CBCL-PR) \cite{Ayub_journal20} for online learning of scenes in this paper. For norm modeling we use the mathematical framework presented in \cite{Sarathy17_IEEE}. Norm learning related to different scene categories is accomplished by employing active learning to ask questions from a human partner. We test our approach on the Pepper robot by taking it through different locations on Penn State University campus for online learning of scenes and their associated norms through question/answer sessions with human partner. 

The remainder of the paper is organized as follows: Section \ref{sec:related_work} reviews the related work regarding scene classification, incremental learning of scenes and norm learning. Section \ref{sec:methodology} describes our complete architecture for online scene and norm learning on a robot. Section \ref{sec:experiments} presents empirical evaluations of the system, demonstrating that our proposed system is capable of learning scenes and norms in an online manner. Finally, Section \ref{sec:conclusion} offers conclusions and directions for future research.

%First talk about norm and the need for scene learning for that. 

%how norms are important for humans and how they learn them in a lifelong manner.
%General norm learning is mostly done for modeling and never really applied on robots because of limitations... models of norms exist.

%However, humans learn scene contexts and the associated norms with that together. Hence, we present a similar system over here. work presented here is a first of its kind.

%We plan to use perception techniques to keep learning scenes and active learning to keep learning norms.

\section{Related Work}
\label{sec:related_work}
Scene classification has been strongly influenced by deep learning \cite{zhou2017,Wang17}. Most approaches, however, have only been tested on indoor scene datasets that do not include natural environments, variations in lighting conditions or the differences stemming from the use of different robots. Most methods also require  the complete dataset, including all of the categories of scenes to be learned, available in a single batch.

For robotics applications, data is available to the robot in the form of a continuous stream. Learning from such streaming video data is known as online learning. Only a few researchers have investigated online learning of scenes for robots. Kawewong et al. \cite{Kawewong13} presents one of the earliest approaches for incremental learning of indoor scenes. They utilize a pre-trained CNN for feature extraction and utilize \textit{n}-value self-organizing and incremental neural networks for incremental learning of scenes. They, however, only test their approach on the MIT 67 scene dataset \cite{Quattoni09} and not on a robot. Paul et al. \cite{Paul_2017_CVPR} presented an active learning approach for online learning of scenes. However, they require the old data to be available when learning new labeled data. Furthermore, they also do not test their approach on a robot or robot-centric indoor scene dataset.

There has been limited work done on norm learning in the context of robotics applications. For example, \cite{Krishnamoorthy18} presents a simple simulated autonomous robot setup in which autonomous cars follow traffic signal norms. Tan et al. \cite{Tan19} present another simple robotics application of learning ownership norms. Neither of these approaches are applicable to the type of context-specific norm learning considered in this paper. Sarathy et al. \cite{Sarathy17_IEEE,Sarathy17} presents a context-specific norm modeling strategy that is based upon Dempster-Shafer theory \cite{shafer_1976}. They present a formal mathematical framework in which norms are modeled using a deontic logic with context-specificity and uncertainty. Unfortunately they do not test their approach on a real robot or with real scene video/image data. In this work, we use the mathematical framework presented in \cite{Sarathy17_IEEE} to model context-specific norms on a Pepper robot in which scene categories (contexts) and related norms are learned in an online manner through active learning.  

\section{Online Scene and Norm Learning}
\label{sec:methodology}

\begin{figure}[t]
\centering
\includegraphics[scale=0.47]{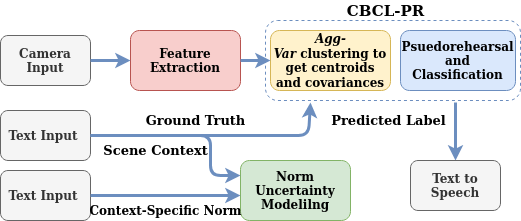}
\caption{\small Overall architecture of our approach. The robot captures new data using its camera and then predicts the label for the scene category using CBCL-PR. The Text-to-Speech module is used to communicate the predicted label. The human provides the true label for the new data to the robot as a text input. The robot further asks the human about the permitted and forbidden norms specific to the scene category.}
\label{fig:online_framework}
\end{figure}

Figure \ref{fig:online_framework} shows the complete architecture of our proposed system. The system can learn new scene categories in an online manner while matching new data to previously learned scene categories and unknown categories. After learning a scene category, the system can learn the norms related to the corresponding scene. The architecture uses Centroid-Based Concept Learning-Pseudorehearsal (CBCL-PR) approach proposed in \cite{Ayub_journal20} to train a robot on new scenes in an online manner with only a single example of a scene available at each increment. For norm modeling and learning, we use the mathematical model proposed in \cite{Sarathy17_IEEE,Sarathy17}. The major components of the architecture are described below.

\subsection{Centroid-Based Concept Learning and Pseudo-Rehearsal(CBCL-PR)}
\label{sec:cbcl}
\noindent CBCL-PR \cite{Ayub_journal20} is a revised version of CBCL \cite{Ayub_2020_CVPR_Workshops} which is a recently developed state-of-the-art method for incremental learning \cite{Ayub_2020_CVPR_Workshops,Ayub_IROS_20} and RGB-D indoor scene classification \cite{Ayub_BMVC20}. CBCL-PR has also been shown to learn objects from only a few examples whereas prior state-of-the-art methods required a large amount of training data, its memory footprint does not grow dramatically as it learns new classes, and its learning time is faster than the other, mostly deep learning, methods. One of the main advantages of CBCL-PR is that it can be adapted to learn from data coming in an online manner such that the new incoming data can belong to already learned classes or completely new classes \cite{Ayub_corl20}. 

\begin{figure}[t]
\centering
\includegraphics[scale=0.36]{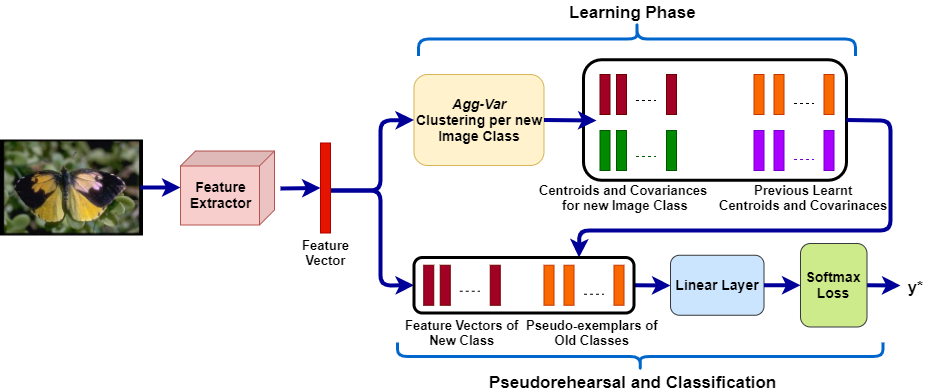}
\caption{\small For each new class of images in a dataset, the feature extractor generates feature vectors for all of the training images in the class. A set of centroids and covariance matrices are created for the feature vectors using the \textit{Agg-Var} clustering algorithm and concatenated with the centroids and covariance matrices of previously learned classes. Pseudo-exemplars for old classes and training examples of new classes are then used to train a shallow network for classification.}
\label{fig:framework_cbcl}
\end{figure}

Figure \ref{fig:framework_cbcl} depicts the architecture for CBCL-PR. It is composed of three modules: 1) a feature extractor, 2) \textit{Agg-Var} clustering, and 3) a pseudo-rehearsal and classification module. In the learning phase, once the human provides the robot with the training examples for a new class, the first step in CBCL-PR is the generation of feature vectors from the images of the new class using a fixed feature extractor. In this paper, we use VGG-16~\cite{Karen14} pre-trained on Places365 dataset~\cite{zhou2017} as the feature extractor. 

\subsubsection{\textit{Agg-Var} Clustering:}
Next, in the case of incremental learning, for each new image class $y$, CBCL-PR clusters all of the training images in the class. \textit{Agg-Var} clustering, a cognitively-inspired approach, begins by creating one centroid from the first image in the training set of class $y$. Next, for each image in the training set of the class, feature vector $x_i^y$ (for the $i$the image) is generated and compared using the Euclidean distance to all the centroids for the class $y$. If the distance of $x_i^y$ to the closest centroid is below a pre-defined distance threshold $D$, the closest centroid is updated by calculating a weighted mean of the centroid and the feature vector $x_i^y$. If the distance between the $i$th image and the closest centroid is greater than the distance threshold $D$, a new centroid is created for class $y$ and equated to the feature vector $x_i^y$ of the $i$th image. 

CBCL also finds the covariance matrices related to each centroid using the feature vectors of all the images clustered in the centroid. The result of this process is a collection containing a set of centroids, $C^y = \{c_1^y, ..., c_{N^*_y}^y\}$, and covariance matrices, $\sum^y = \{\sigma_1^y,...,\sigma_{N^*_y}^y\}$, for the class $y$ where $N^*_y$ is the number of centroids for class $y$. This process is applied to the sample set $X^y$ of each class incrementally once they become available to get a collection of centroids $C  = C^1, C^2, ..., C^N$ and covariance matrices $\sum = \sum^1, \sum^2, ..., \sum^N$ for all $N$ classes in a dataset. %It should be noted that using the same distance threshold for different classes can yield a different number of centroids per class depending on the similarity among the images in each class. Hence, one only needs to tune a single parameter ($D$) to get the optimal number of centroids (in terms of validation accuracy) in each class. 
Note that \textit{Agg-Var} clustering calculates the centroids for each class separately. Thus, its performance is not strongly impacted by when the classes are presented incrementally. 

\subsubsection{Pseudorehearsal:}

For classification of images from all the classes seen so far, CBCL-PR trains a shallow neural network composed of a linear layer trained with a softmax loss on examples from the current increment and pseudo-exemplars of the old classes (pseudorehearsal). Pseudo-exemplars are not real exemplars of a class, rather they are generated based on the class statistics to best resemble the actual exemplars. 

CBCL-PR uses each of the centroids $c_i^y$ and covariance matrices $\sigma_i^y$ of class $y$ to generate a set of pseudo-exemplars. First a multi-variate Gaussian distribution is created using the centroid as the mean and the corresponding covariance matrix. The Gaussian distribution is then sampled to generate the same number of pseudo-exemplars as the total number of exemplars represented by the centroid/covariance matrix pair $c_i^y$,$\sigma_i^y$. This process leads to a total of $N_y$ number of pseudo-exemplars for class $y$. In this way the network is trained on $N_y$ number of examples for each class $y$ in each increment. For classification of a test image, the feature vector $x$ of the image is first generated using the feature extractor. Next, this feature vector is passed through the shallow network to produce a predicted label $y^*$. A more detailed description of CBCL-PR can be found in \cite{Ayub_2020_CVPR_Workshops,Ayub_BMVC20,Ayub_journal20}

\subsection{Predicting Unknown Scene Categories}
CBCL-PR is designed for incremental learning where the test images always belong to one of the learned categories. However, for the online learning setting considered in this paper the robot also has to make a prediction about an unknown scene category. Further, in \cite{Ayub_journal20} CBCL-PR is provided with the complete training set of a class at one time, however for the online learning scenario considered here, data belonging to a previously known scene category can also become available in an increment.

For this work we have adapted CBCL-PR to operate in an online learning manner. First, to make a prediction about an unknown scene category, we calculate the Euclidean distance of the feature vector of the new scene image to all of the learned centroids. If the distance is below the distance threshold $D$ that was used to get the centroids, we predict that the new image belongs to a new scene category. Second, if the ground truth provided by the human for a new scene image belongs to one of the already learned categories, we simply apply \textit{Agg-Var} clustering on the new scene images to either update the previously learned centroids and covariance matrices of the scene category or create new centroids and covariance matrices based upon the distance threshold $D$. A more detailed explanation for adapting CBCL-PR for online learning is provided in \cite{Ayub_corl20}.

\subsection{Norm Modeling and Learning}
We follow the norm modeling framework presented in \cite{Sarathy17_IEEE,Sarathy17}. In particular, we use the idea of a context-specific belief-theoretic norm which is of the form:

\begin{equation}
    \mathcal{N} \myeq \mathbb{D}_C^{[\alpha,\beta]}A
\end{equation}

\noindent where $\mathcal{N}$ is a context-specific belief theoretic norm for a formal language $\mathcal{L}$, $\mathbb{D}_C^{[\alpha,\beta]}$ is an uncertain context-specific deontic operator
with $[\alpha,\beta]$ representing a Dempster-Shafer uncertainty interval \cite{shafer_1976} for the operator with $0<\alpha<\beta<1$. $\mathbb{D}$ is a collection of three deontic model operators: obligatory $\mathbb{O}$, forbidden $\mathbb{F}$ and permissible $\mathbb{P}$. $C\in \mathcal{L}$ represents the context and $A\in \mathcal{L}$ represents a state or an action. The above mentioned norm expression states that in a context $C$, an action or state $A$ is either obligatory, forbidden or permissible or not obligatory, forbidden or permissible. 

We use the mathematical model of equation (1) to model and learn norms for different scene categories (contexts) in an online manner. The initial knowledge of the robot consists of a set of possible actions that the robot can perform. For example, \textit{talk} is an action available to the robot where it can use its speech module to communicate. However, the robot does not know which of its known actions are permissible, obligatory or forbidden in a particular context. Hence, for each new context the robot learns about context-specific norms through active learning i.e. by having a short question/answer session with a human partner. We only use yes/no question types in this paper. In each new context/scene, the robot asks its human partner which of its actions are obligatory, forbidden or permissible. If a norm does not exist in the robot's knowledge base, it simply creates a new norm for a particular context. The new norm is initialized with an uncertainty interval of either $[0,0]$ or $[1,1]$ based upon the answer given by the human (yes or no). However, if a norm already exists in the robot's knowledge base and it gets an answer from its human partner, it updates the uncertainty interval of the norm based on the new answer. In section \ref{sec:experiments} we present results of learning different context-specific norms by the Pepper robot. The learned norms and the associated uncertainties can be used by the robot to reason about before performing an action in a given context. Experiments regarding norm reasoning for performing actions are left for future work.

\section{Experiments}
\label{sec:experiments}

\begin{figure}[t]
\centering
\includegraphics[scale=0.6]{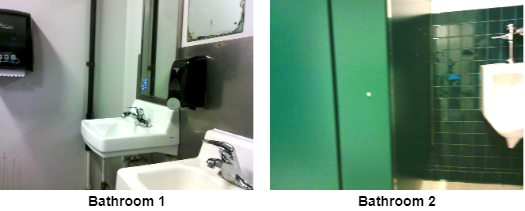}
\caption{\small Images from two different bathrooms collected by the Pepper robot. Both images are drastically different from each other. }
\label{fig:scenes}
\end{figure}

We evaluate our method for online learning of scenes and related norms on the Pepper robot.
The robot was manually driven through 10 different locations belonging to 5 different categories (bathroom, classroom, library, Office, kitchen) on the Penn State University campus. The locations differed from one another even when belonging to the same category, making the learning and recognition task difficult. For example, Figure \ref{fig:scenes} shows images from two different bathrooms captured by the Pepper robot. Note how different the images are.  Images were captured using the front head camera of the Pepper with a resolution of 320$\times$240. Pepper's text-to-speech module was used to communicate with the human while the labels and norm related answers were provided by the human using a keyboard.

\subsection{Online Scene Learning Using the Pepper Robot}
Our first experiment demonstrates our method for online scene learning on a Pepper robot. For each new location, the robot took the input of a scene (either belonging to earlier learned categories or a new category) using its camera. Next, it used the method described in Figure \ref{fig:online_framework} to either predict the label of the scene or predict that the scene category was unknown. The robot further asked the human to confirm the label of the new scene. If the human provided the label, the text module passed it to CBCL-PR to update the previous centroids, otherwise if the new scene belonged to a known category the robot created new centroids for the new scene category. 

\begin{figure}[t]
\centering
\includegraphics[scale=0.5]{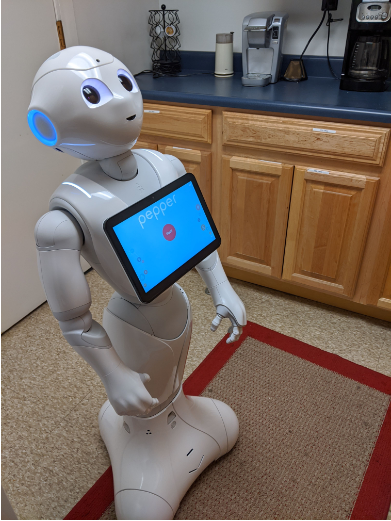}
\caption{\small An example of the Pepper robot trying to predict the category \textit{kitchen}.}
\label{fig:pepper}
\end{figure}

It took approximately 30 seconds to capture a video, while feature extraction took about 11 seconds. The time required to make a prediction about the new video was only 1 second, while confirmation of the label from the human took approximately 5 seconds. \textit{Agg-Var} clustering took only 1 second to find the centroids and pseudorehearsal and classifier training took 10 seconds. Hence, the total time to learn about a new location was approximately 58 seconds. It should be noted that after receiving the label from the human, Pepper continue to process the data while moving to other locations, so it only needs 47 seconds at one location to make a prediction.

At each of the locations, Pepper captured the video data by rotating its head from center to the left and then left to right for 20 seconds. After capturing the video, the feature extractor was used to extract the features for all the images. Our method (Figure \ref{fig:online_framework}) used the shallow classifier in CBCL-PR to make a prediction about the category of the location, while it used the distance to the closest centroid to make a prediction if the location was known. Pepper's text-to-speech module was then used to communicate the prediction. After making the prediction, Pepper confirmed its prediction in the case of a known category or asked for a new label in the case of an unknown category. The experimenter then typed the ground truth of the location in a terminal. Pepper repeated the typed label to confirm with the experimenter. After the confirmation, \textit{Agg-Var} clustering in CBCL-PR was used to find the centroids for the new data in the case of an unknown category or update the previously learned centroids in the case of a known category. After learning the centroids that represent the new location, the image data was discarded to satisfy the conditions of online learning. Figure \ref{fig:pepper} shows the Pepper robot in a kitchen trying to predict the location's label.

Of the 10 locations tested, Pepper was presented with 5 locations that belonged to new categories and 5 locations from a previously known category. Pepper correctly predicted 80\% of locations when the location belonged to a new category and 80\% of locations when predicting known categories. These results clearly show the effectiveness of our approach to mitigate \textit{catastrophic forgetting} when learning scene categories using streaming video data on a robot. Where catastrophic forgetting is a problem in continual learning in which the model forgets the previously learned classes when learning new classes and the overall accuracy decreases drastically. Detailed experiments on incremental learning on benchmark datasets for CBCL-PR are available in \cite{Ayub_2020_CVPR_Workshops,Ayub_journal20}. %These results clearly show the effectiveness of our approach for learning scene categories using streaming video data on a robot, which can lead to further novel applications like higher-level localization of a robot in unknown buildings.

\begin{table}[t]
\centering
%\small
\begin{tabular}{ |P{2.5cm}|P{7cm}| }
     \hline
    \textbf{Robot Actions} & \textbf{Description} \\
     \hline
     \textit{talkLoundly} & Speak in a higher volume \\
     \hline
     \textit{talkQuietly} & Lower volume when speaking\\
     \hline
     \textit{beQuiet} & Do not speak\\
     \hline
     \textit{listen} & Process the incoming audio input\\
     \hline
     \textit{watch} & Capture images through camera for processing\\
     \hline
     \textit{walk} & Move around\\
 \hline
 \end{tabular}
 \bigskip
 \caption{Description of the six action types available to the Pepper robot.}
 \label{tab:actions}
 \end{table}

\subsection{Norm Learning by the Pepper Robot}
The setup for this experiment was the same as the scene learning experiment. For each new scene, after the robot predicted the scene and got the ground truth label from the human partner, it initiated a question/answer session regarding norm learning. For each location, Pepper was allowed to ask at most three questions. In the three questions, Pepper randomly chose which norms to ask about. The limit of three questions was set arbitrarily, keeping in mind that the human would get annoyed if he/she has to answer a large number of questions at a location. All of the questions were related to the permission norms i.e. if one of the robot actions was permissible at a location. The initial set of actions available to Pepper in its knowledge base were: \textit{talkLoudly}, \textit{talkQuietly}, \textit{beQuiet}, \textit{listen}, \textit{watch} and \textit{walk}. Table \ref{tab:actions} provides a description of the six actions. Pepper used its speech to text module to ask questions about the norms. The experimenter answered the questions using a keyboard. The experimenter's answers were based on the results shown by previous studies in norm learning \cite{Sarathy17_IEEE,Sarathy17}. Using the answers provided by the experimenter, Pepper updated the uncertainties related to the previously learned context-specific norms and also created new context-specific norms with 100\% certainty. 

Table \ref{tab:permitted_norms} shows all the permission norms learned by Pepper at the 10 locations (5 scene categories). Most of the action types have either 100\% or 0\% uncertainties. This is because of the limited number of actions available to the robot and the limited amount of locations visited per scene category (2 per scene category). In cases when the experimenter gave different answers in different locations of a scene category about an action, the uncertainties were updated. For example, for the action type \textit{talkQuitely} in the library context the agent said that it was not permissible in the first library. Hence, Pepper's initial uncertainty interval for this action was [0,0]. However, in the second library, the experimenter said \textit{talkQuietly} was permissible. Therefore, Pepper updated the uncertainty interval to [0,0.5]. The robot was not able to ask questions about some actions at all because of the limited number of actions allowed to be asked. The empty entries in Table \ref{tab:permitted_norms} show that the robot did not ask about those actions at the corresponding location. For example, the robot never asked if action \textit{watch} was permissible in the library.

The time required to communicate with the human partner for three questions was about 10 seconds. These results depict the ability of the robot to learn context-specific norms in an online manner without forgetting any previously learned information. Social robots can use these context-specific norms while reasoning about which actions to perform in a particular context. The norm reasoning module is left for future work. 

\begin{table}[t]
\centering
%\small
\begin{tabular}{|P{2cm}|P{1.8cm}|P{1.8cm}|P{1.5cm}|P{1.5cm}|P{1.5cm}| }
     \hline
    \textbf{Permitted Action} & \textbf{Bathroom} & \textbf{Classroom} & \textbf{Library} & \textbf{Office} & \textbf{Kitchen} \\
     \hline
     \textit{talkLoudly} & [0,0] & [0,0] & [0,0] & [0,0] & [1,1]\\
     \hline
     \textit{talkQuietly} & [1,1] & [0,0.5] & [0,0.5] & [1,1] & [0.5,1]\\
     \hline
     \textit{beQuiet} & [1,1] & [0,0.5] & [0.5,1] & [0,0.5] & [1,1]\\
    \hline
     \textit{listen} & [1,1] & [1,1] & - & [1,1] & [1,1]\\
     \hline
     \textit{watch} & [0,0] & - & [1,1] & - & [1,1]\\
     \hline
     \textit{walk} & [1,1] & - & - & [1,1] & - \\
 \hline
 \end{tabular}
 \bigskip
 \caption{Permission norms with uncertainty intervals learned by Pepper for five different scene categories.}
 \label{tab:permitted_norms}
 \end{table}

%for norm learning, say that for each location, the agent was allowed to ask only one question per three norm conditions. 

%Actions available to the robot were: talk loud, talk quietly, no talk, shut off camera/don't watch. 

%pepper has to learn the norms and the uncertainties. answers were provided by the experimenter. 

%make a table on what the pepper learned over time. 

%experiments on pepper, FIRST only add that and add the dataset stuff if there is room.

%For pepper experiment, you show how the scenes are learnt, and then how the norms are updated with different answers in different scenes, just like the original paper. 
%Robot can ask 9 norm specific questions in one session. i.e. it can confirm about 3  behaviours/norms for each of the three types. 
\section{Conclusion}
\label{sec:conclusion}
This paper presents and evaluates a method for a novel problem: online learning of scenes and context-specific norms on a robot in unconstrained environments. We evaluate our approach on the Pepper robot to learn various scene categories and related norms in an online manner. 
The work presented here makes only a few assumptions. We assume that the human provides correct labels and correct answers related to norms when asked by the robot. Incorrect information would certainly negatively impact the performance of our approach.

Nevertheless, we believe that this paper makes several important contributions. Most importantly, we offer a novel and realistic approach to online scene and context-specific norm learning. This method allows the robots to not only learn different categories and appropriate behaviors associated with these categories but to also use the recognition of these different environments to moderate their behavior and decision-making. For example, allowing a robot to recognize that its current location is in a library and that it should therefore reduce the volume of its speech. Our future work will focus on the reasoning module to choose appropriate actions based on learned context-specific norms. Our overarching hope is that this work will evolve into new applications and competencies for a variety of robot applications.  

\section*{Acknowledgements}
This work was partially supported by the Air Force Office of Sponsored Research contract FA9550-17-1-0017 and National Science Foundation grant CNS-1830390. Any opinions, findings, and conclusions or recommendations expressed in this material are those of the author(s) and do not necessarily reflect the views of the National Science Foundation.

\bibliographystyle{unsrt}
\bibliography{main}%,IROS.bib,CVPR.bib,CVPR_scenes.bib}

\end{document}